\documentclass[review]{elsarticle}

\usepackage{geometry}
\usepackage{tabularx}
\usepackage{booktabs}
\usepackage{caption}
\usepackage{float}
\usepackage{amsmath}
\usepackage{bm}
\usepackage{xcolor}
\usepackage{color,soul}
\usepackage{tabu}
\usepackage{textcomp}
\usepackage[normalem]{ulem}
\usepackage{amsmath}
\usepackage{multirow}
\usepackage{lscape}
\newcommand{\stkout}[1]{\ifmmode\text{\sout{\ensuremath{#1}}}\else\sout{#1}\fi}

\makeatletter 
\newcount\SOUL@minus
\makeatother  

\journal{ISPRS Journal of Photogrammetry and Remote Sensing}

\biboptions{authoryear}

\begin{document}

\clearpage
\setcounter{page}{1}

\begin{frontmatter}

\title{Plant Species Richness Prediction from DESIS Hyperspectral Data: A Comparison Study on Feature Extraction Procedures \\ and Regression Models}

\author[address1]{Yiqing Guo}
\author[address1]{Karel Mokany}
\author[address2]{Cindy Ong}
\author[address3]{Peyman Moghadam}
\author[address1]{Simon Ferrier}
\author[address4]{Shaun R. Levick}

\address[address1]{CSIRO Land and Water, Acton, ACT 2601, Australia}
\address[address2]{CSIRO Energy, Kensington, WA 6151, Australia}
\address[address3]{CSIRO Data61, Pullenvale, QLD 4069, Australia}
\address[address4]{CSIRO Land and Water, Winnellie, NT 0822, Australia}

\begin{abstract}
The diversity of terrestrial vascular plants plays a key role in maintaining the stability and productivity of ecosystems. Monitoring species compositional diversity across large spatial scales is challenging and time consuming. Airborne hyperspectral imaging has shown promise for measuring plant diversity remotely, but to operationalise these efforts over large regions we need to advance satellite-based alternatives. The advanced spectral and spatial specification of the recently launched DESIS (the DLR Earth Sensing Imaging Spectrometer) instrument provides a unique opportunity to test the potential for monitoring plant species diversity with spaceborne hyperspectral data. This study provides a quantitative assessment on the ability of DESIS hyperspectral data for predicting plant species richness in two different habitat types in southeast Australia. Spectral features were first extracted from the DESIS spectra, then regressed against on-ground estimates of plant species richness, with a two-fold cross validation scheme to assess the predictive performance. We tested and compared the effectiveness of Principal Component Analysis (PCA), Canonical Correlation Analysis (CCA), and Partial Least Squares analysis (PLS) for feature extraction, and Kernel Ridge Regression (KRR), Gaussian Process Regression (GPR), and Random Forest Regression (RFR) for species richness prediction. The best prediction results were $r=0.76$ and $\text{RMSE}=5.89$ for the Southern Tablelands region, and $r=0.68$ and $\text{RMSE}=5.95$ for the Snowy Mountains region. Relative importance analysis for the DESIS spectral bands showed that the red-edge, red, and blue spectral regions were more important for predicting plant species richness than the green bands and the near-infrared bands beyond red-edge. We also found that the DESIS hyperspectral data performed better than Sentinel-2 multispectral data in the prediction of plant species richness. Our results provide a quantitative reference for future studies exploring the potential of spaceborne hyperspectral data for plant biodiversity mapping. \\ \end{abstract}


\end{frontmatter}


\section{Introduction}

Plant biodiversity is of critical importance to the stability of terrestrial ecosystems \citep{frankel1995conservation}. Anthropogenic activities, such as inappropriate cropping, deforestation, overgrazing, and construction, in conjunction with climate change, have been leading to substantial degradation and loss of natural habitats, posing imminent threats to vulnerable plant species \citep{ceballos2015accelerated, tollefson2019humans}. Consequently, species extinctions are occurring much faster than the natural background rate \citep{ceballos2015accelerated}. Conservation activities have been undertaken in many places around the world aiming to reduce the current rate of extinction \citep{leclere2020bending, mokany2020reconciling}.

Spatial mapping of plant biodiversity helps with a better understanding of the distribution and temporal trends of plant species richness, facilitating effective policy making in environmental conservation and restoration \citep{stevenson2021matching, myers2021new, de2021annual}. Considerable effort has been made to collect samples of plant species richness through in-situ surveys \citep{kattge2020try}. Despite the ever increasing amount of data, it has been recognised that the completeness and representativeness of biodiversity samples still remain as a major challenge for the compilation of up-to-date biodiversity maps with fine resolution and wide coverage \citep{konig2019biodiversity, kattge2020try}. This data gap is understandable as field expeditions are labour and time consuming, and sometimes infeasible if the location is remote or hard to access \citep{wang2019remote, guo2018effective}. Moreover, inconsistencies among collection campaigns in their sampling strategies and ground plot sizes, confounded by human subjectivity and bias, further hampered the use of ground sampling data in downstream scientific research \citep{wang2019remote}.

Spaceborne remote sensing has long been deemed as a promising and cost-effective tool for mapping plant biodiversity, mainly due to its ability to capture data over large areas and in a timely manner \citep{skidmore2021priority, wang2019remote, bush2017connecting, pettorelli2016framing}. Among different types of remote sensing data, hyperspectral data is of particular interest for the biodiversity community, as it contains rich features in the spectral domain that can be utilised to explore the underlying relationship with plant biodiversity on the ground \citep{ghiyamat2010review, carlson2007hyperspectral, guo2022quantitative}. Previous studies have shown that plant biodiversity is linked to remotely sensed spectral measurements because of a well-founded interrelationship between plant species richness and primary productivity \citep{wang2016seasonal, grace2016integrative}. It is hypothesised that a high diversity of plant species enhances primary production of the community, as a result of complementary functions provided by the diversified species composition. These interrelationships have the potential to enable researchers to use hyperspectral measurements, and their derived spectral indices, as remotely sensed measures of vegetation productivity and estimates of species richness.

Most research into the relationships between biodiversity and hyperspectral data have made use of hand-held or airborne hyperspectral sensors, which are more readily available than spaceborne hyperspectral data-streams (e.g., \cite{peng2018assessment}, \cite{asner2008hyperspectral}, \cite{jeanbaptiste2014}, and \cite{hacker2020retrieving}). Spaceborne hyperspectral imaging is relatively rare, with no active satellites in orbit since the Hyperion mission (which was active from 2000 to 2017). However, in preparation for the upcoming launch of EnMAP \citep{guanter2015enmap}, the DLR launched an exploratory system to the International Space Station and embedded into the Multi-User-System for Earth Sensing (MUSES) platform in 2018. The DLR Earth Sensing Imaging Spectrometer (DESIS) \citep{eckardt2015desis, mafanya2022assessment} provides a unique opportunity to test the potential for monitoring plant species diversity with spaceborne hyperspectral data. It delivers hyperspectral images with 235 spectral bands over the visible and near-infrared regions of 400 $\sim$ 1000 nm, with a spectral resolution of 2.55 nm and a spatial resolution of 30 m \citep{eckardt2015desis, alonso2019data, krutz2019instrument}. The high resolutions in both spectral and spatial domains make DESIS a promising data source for estimating biodiversity from space. However, there is so far a lack of quantitative studies on assessing the ability of DESIS hyperspectral data for predicting plant species richness values on the ground. It is worth noting the added challenge of spaceborne hyperspectral imagery such as DESIS having a larger pixel size (30m) than typical airborne (ranging from a few centimeters to several meters). As the signal of a large pixel tends to comprise a mix of multiple species, many techniques previously used to quantify richness with hyperspectral data, which detect individual species and then sum up to get richness (e.g., \cite{asner2008hyperspectral,jeanbaptiste2014}), cannot be applied.

The original bands in hyperspectral imagery are not orthogonal but rather highly collinear with each other, presenting a high degree of redundancy of information. Such redundancy can be removed to some extent by transforming the original spectral bands into an orthogonal space of a lower dimensionality. Among popular algorithms for dimensionality reduction are Principal Component Analysis (PCA) \citep{xu2019thin, jia1999segmented}, Canonical Correlation Analysis (CCA) \citep{zhao2014earlya, richards1999remote}, and Partial Least Squares analysis (PLS) \citep{feilhauer2015multi, hacker2020retrieving, wang2019mapping}. These dimensionality reduction techniques reduce information redundancy by removing multicollinearity among spectral bands, enabling extracting spectral features from the original spectral data of hundreds of bands. In contrast to feature selection that chooses a subset of the original bands (or spectral indices computed from a subset of the original bands), feature extraction techniques such as PCA, CCA, and PLS are able to makes use of information in all bands by transforming them into compact yet informative features. Compared with pre-defined vegetation indices such as Ratio Vegetation Index (RVI) and Normalised Difference Vegetation Index (NDVI), spectral features generated with feature extraction have shown better performance in extracting useful information from hyperspectral measurements \citep{zhao2014earlya}. Following the extraction of spectral features, regression analysis can then be conducted to explore potential relationships between the extracted features and target biological variables. Commonly applied regression algorithms include the Kernel Ridge Regression (KRR), Gaussian Process Regression (GPR), and Random Forest Regression (RFR). Statistical regression based on extracted spectral features has shown to be effective in addressing biological problems with hyperspectral remote sensing data. For example, in a study to detect unintended herbicide damage in crops with hyperspectral measurements, dimensionality reduction was conducted with CCA in order to extract useful information to discriminate between healthy and damaged crops \citep{zhao2014earlya}. In another study aiming to retrieve foliar traits from hyperspectral data, PLS was used to reduce the spectral dimensionality, with adequate retrieval accuracies being achieved for 10 out of the 11 functional traits \citep{hacker2020retrieving}. These studies demonstrate that feature extraction and statistical regression can be effective tools in addressing biological problems with hyperspectral measurements. 

In this study, we aimed to assess the potential for DESIS hyperspectral data to predict on-ground plant species richness in two regions of southeast New South Wales, Australia---the Southern Tablelands and Snowy Mountains. Our approach focused on spectral feature extraction from the DESIS spectra, and subsequent regression against field-measured plant species richness. We tested the combination of different feature extraction procedures (PCA, CCA, and PLS) and regression models (GPR, KRR, and RFR) for the predictive performance of species richness with DESIS data. Through quantitative analyses, we sought to address primarily the following important questions: (1) How much variation in plant species richness can be explained with DESIS data? (2) Which parts of the spectrum had the most explanatory power? (3) Could similar results be achieved with more readily available multi-spectral imagery such as Sentinel-2?

\section{Materials and Methods}

While DESIS hyperspectral data contain rich spectral information, modelling is needed to link such information to field-based measurements of plant species richness. Here we followed a two-step approach whereby spectral features were first extracted from DESIS spectra and then correlated to species richness through regression. In this section, we start with describing the DESIS spectra and in-situ richness samples, followed by introducing the methods for feature extraction, regression, and accuracy assessment.

\subsection{Study site and field data}

\begin{figure}[htb!]
\centering
\captionsetup{font=normalsize}
\includegraphics[width=9 cm]{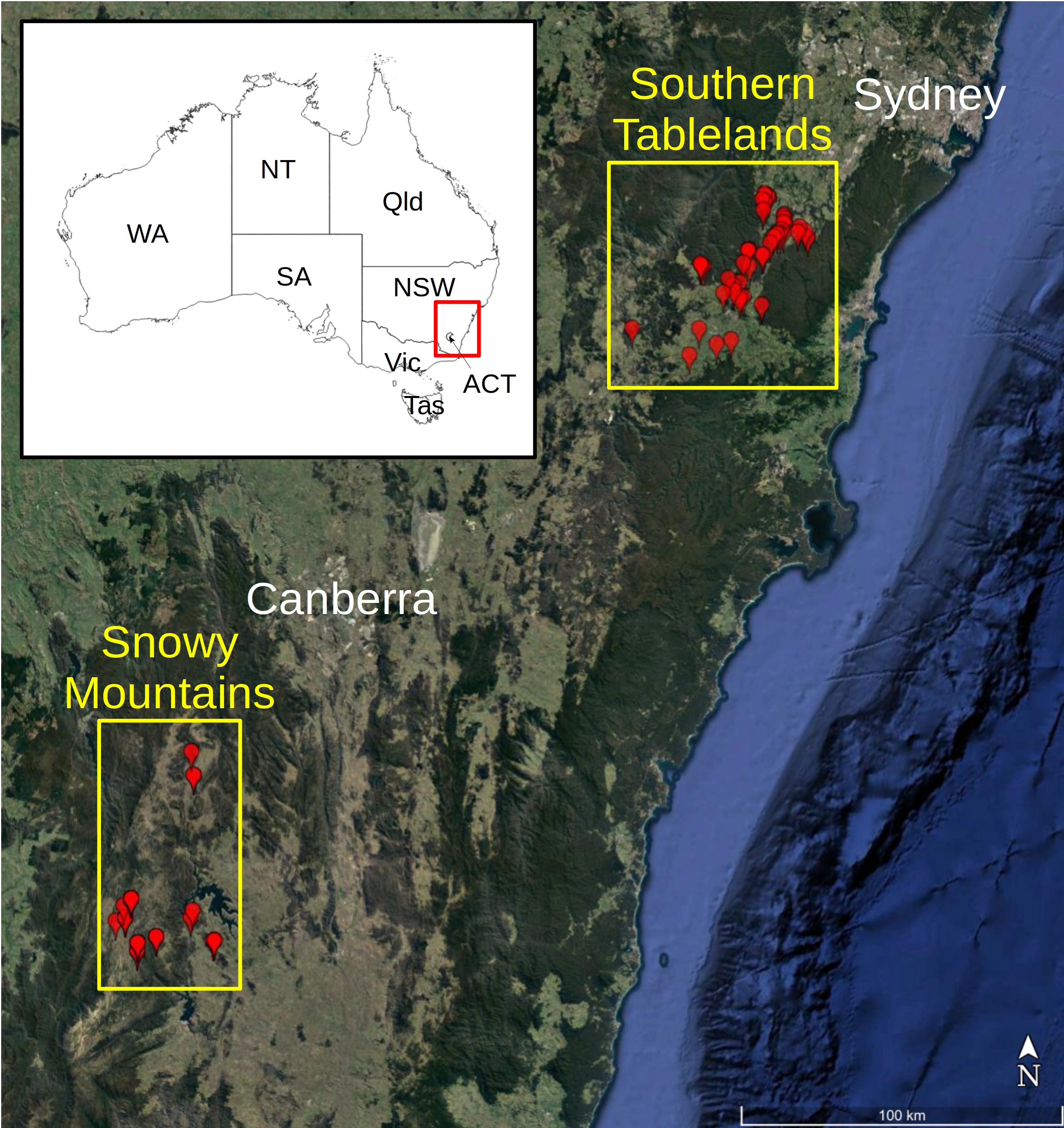}
\caption{Locations of in-situ plant species richness samples collected in field experiments.\label{fig:richness}}
\end{figure}

This study focused on two different habitat types in southeast New South Wales, Australia, namely the Southern Tablelands (34°12'26''--34°39'07''S, 150°05'57''--150°40'51''E) and Snowy Mountains (35°43'58''--36°16'30''S, 148°23'16''--148°39'02''E), as shown in Fig. \ref{fig:richness}. Both regions are located within the climate zone of Cfb (oceanic climates), according to the Köppen–Geiger climate classification system.

The Southern Tablelands region is located to the southwest of Sydney (Fig. \ref{fig:richness}). It is characterised by high-altitude plains with a rich biodiversity. There are more than 1200 plant species within Southern Tablelands, of which 30 are listed as threatened \citep{fallding2002planning}. A large part of the landscape has been transformed into suburbs for residential developments and pastures for grazing purposes. Considering the high degree of human interference and habitat alteration, conservation efforts have been undertaken in order to preserve endangered plant species by improving habitat connectivity and condition. 

The Snowy Mountains region is located to the southwest of Canberra (Fig. \ref{fig:richness}). It encompasses the highest mountain ranges of the Australian Alps,  serving as an important habitat for alpine-exclusive species. There are 212 species of vascular plants, of which 21 are endemic \citep{pickering2008vascular}. Due to its unique status in Australia's ecosystem, the plant biodiversity in Snowy Mountains has drawn consistent interest from the research community (e.g. \cite{korner1995alpine}, \cite{pickering2008vascular}, and \cite{pickering2009vascular}).

For on-ground measures of vascular plant species richness, we obtained plant community survey data from the NSW BioNet Vegetation Information System database \citep{nsw2019bionet}. Field surveys were conducted to collect species richness samples in 2016 and 2017 for the two regions, with a sampling plot area of 400 m$^2$ (20 m $\times$ 20 m) at each surveying location. A significant bushfire event occurred during the 2019--2020 summer, with some of the sampling points situated within the affected areas. These bushfire-affected samples were excluded from the data set, based on the National Indicative Aggregated Fire Extent Datasets (NIAFED) provided by the Australian Government Department of Agriculture, Water and the Environment. After the exclusion, a total of 44 and 29 samples were used in this study for analysis for the Southern Tablelands and Snowy Mountains regions, respectively. The locations of these samples are shown in Fig. \ref{fig:richness}, and their associated information is summarised in Table \ref{tab:samples}. For each sampling plot, the number of native vascular plant species was calculated and used as the response variable in our analyses. 

\begin{table}[htb!] 
\captionsetup{font=normalsize}
\caption{Information summary of the plant species richness samples collected in field observations.\label{tab:samples}}
\begin{tabularx}{\textwidth}{l l l}
\toprule
\textbf{}	& \textbf{Southern Tablelands}	& \textbf{Snowy Mountains}\\
\midrule
Number of Samples	 & 44 & 29\\
Sampling Time		& Feb 19, 2017 $\sim$ Dec 07, 2017 & Feb 24, 2016 $\sim$ Dec 13, 2017\\
Plot Area		& 400 m$^2$ (20 m $\times$ 20 m) & 400 m$^2$ (20 m $\times$ 20 m)\\
Geo-extent		& \begin{tabular}[c]{@{}l@{}}34°12'26''--34°39'07''S\\ 150°05'57''--150°40'51''E\end{tabular} & \begin{tabular}[c]{@{}l@{}}35°43'58''--36°16'30''S\\ 148°23'16''--148°39'02''E\end{tabular}\\
\bottomrule
\end{tabularx}
\end{table}

\begin{figure}[htb!]
\centering
\captionsetup{font=normalsize}
\includegraphics[width=11cm]{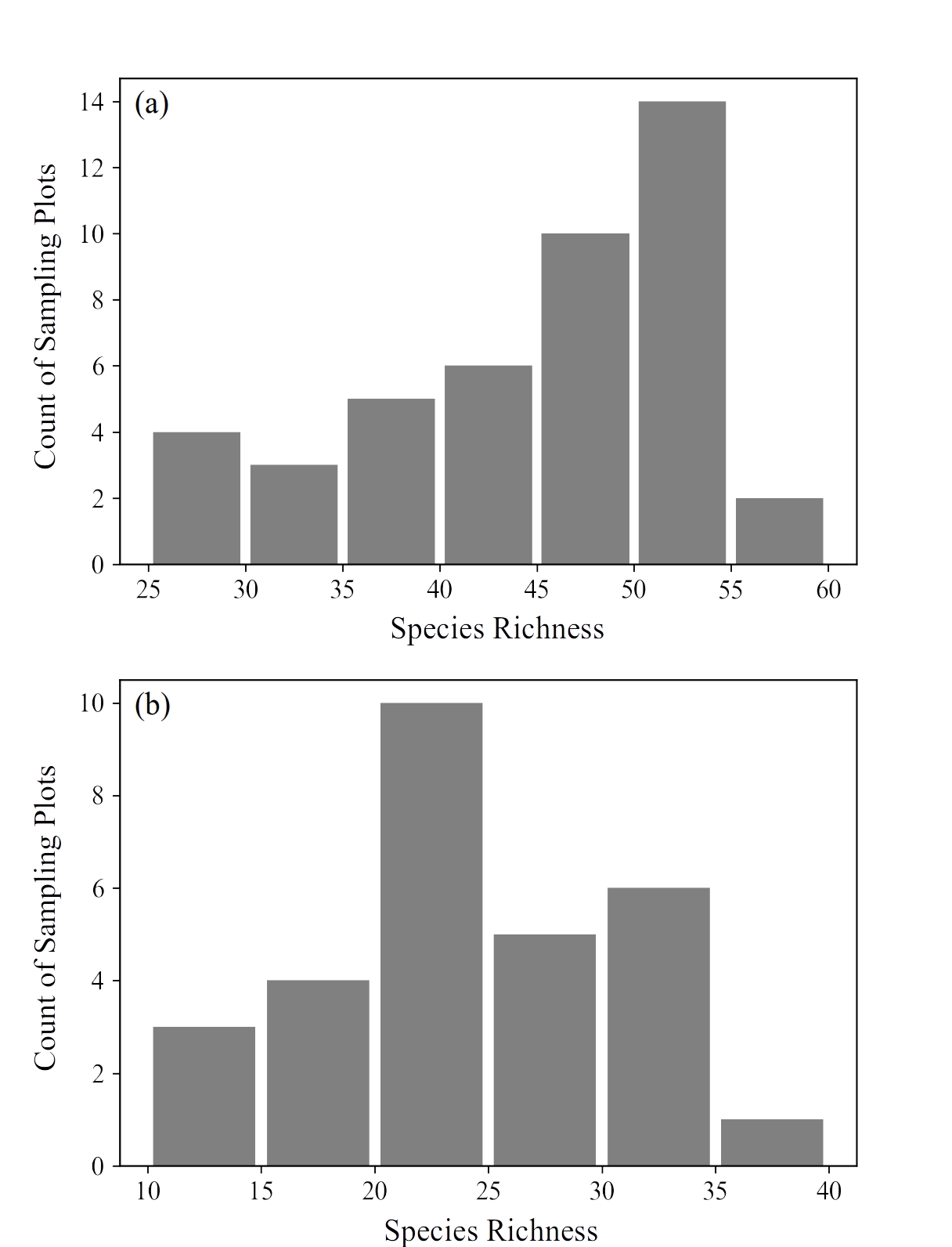}
\caption{Histograms of species richness distribution for sampling plots in the (a) Southern Tablelands, and (b) Snowy Mountains regions.\label{fig:hist}}
\end{figure}

The histograms of species richness distribution for sampling plots in the two regions are shown in Fig. \ref{fig:hist}. Generally, the sampling plots in Southern Tablelands show a higher richness of species than Snowy Mountains, with a mean richness value of 44.4 for the former and 23.5 for the latter. The difference in species richness can be mainly attributed to the fact that the Southern Tablelands is located at lower altitudes with a relatively warmer climate than the mountainous region of Snowy Mountains.

\subsection{Satellite Data}\label{ssec:satellite_data}

The DESIS spectrometer \citep{krutz2019instrument} is embedded in the MUSES platform onboard the International Space Station at an altitude of approximately 400 km. It operates in a push-broom imaging mode featuring state-of-the-art radiometric and spectral specifications. It delivers hyperspectral images with 235 spectral bands over the visible and near-infrared regions of 400 $\sim$ 1000 nm, with a spectral resolution of 2.55 nm and a spatial resolution of 30 m. The radiometric resolution for each band is 12 bit with 1 bit gain. The signal-to-noise ratio is 195 at the wavelength of 550 nm.

In our study, the DESIS Level-2A product was used. It consisted of surface reflectance images with atmospheric correction having been applied. The correction was conducted with DLR's PACO (Python Atmospheric COrrection) software \citep{de2018validation} where the MODTRAN\textsuperscript{\textregistered} radiative transfer model \citep{berkmodtran} served as the module for simulating atmospheric effects. As inputs for atmospheric simulation, aerosol optical thickness and water vapour content were retrieved per pixel using reflectance in the red and NIR bands, and bands around the water absorption features of 820 nm, respectively. DESIS spectra intersecting with locations of the species richness samples were queried within CSIRO's Earth Analytics and Science Innovation (EASI) platform. We selected spectra captured in January 2020 for our analysis. In order to moderate the random noise present in the original spectra, the spectral resolution were down-sampled from 2.55 nm into 10.2 nm bins with the assumption of a Gaussian-shaped spectral response function. The atmospherically affected bands of 759, 769, 933.4, 943.4, and 953.2 nm, and the low quality bands of 402.8, 410.3, and 999.5 nm at the left and right ends of the spectrum were removed. A total of 52 bands were retained after the removal. Pixels flagged as cloud by the DLR Level-1 processing were masked out. 

The Bidirectional Reflectance Distribution Function (BRDF) effect in DESIS data is unneglectable, given the $4.1^{\circ}$ field of view in conjunction with the $\pm15^{\circ}$ along track pointing capability of the DESIS sensor, and the $\pm25^{\circ}$ along track and $-45^{\circ}$ $\sim$ $+5^{\circ}$ cross track tilting capability of the MUSES platform. Following the approach adopted in \cite{green1985analysis} and \cite{ong2014mapping} for correcting BRDF effect in hyperspectral data, in our study each DESIS spectra was mean-normalised with each spectral band being divided by the mean value over all bands.

\begin{figure}[htb!]
\centering
\captionsetup{font=normalsize}
\includegraphics[width=14.5 cm]{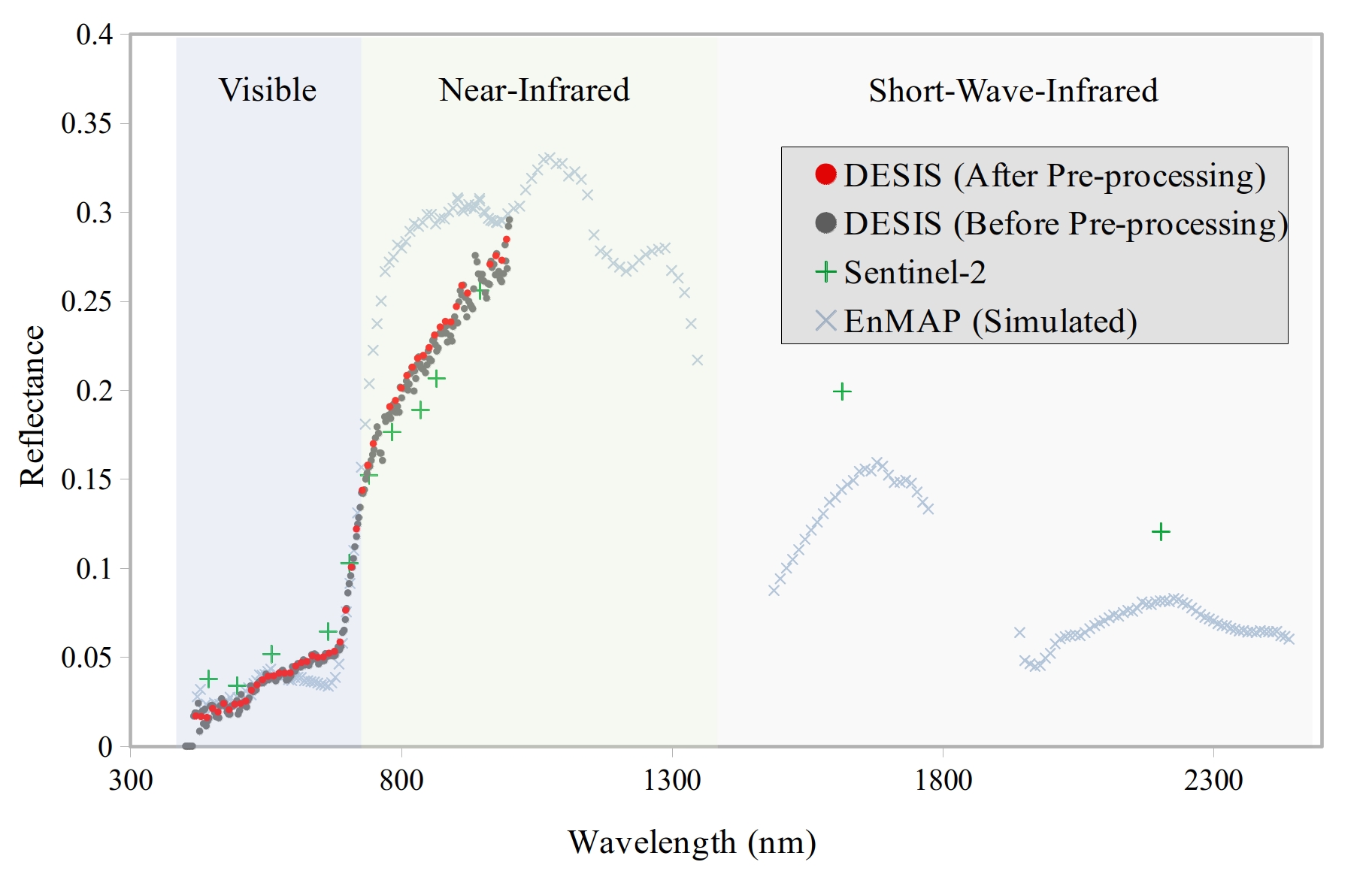}
\caption{Comparison of the DESIS (before and after pre-processing) and Sentinel-2 spectra at one of the ground sampling plots. An EnMAP spectrum simulated for a random location is also shown for reference.\label{fig:compare}}
\end{figure}

For comparison purpose, cloud-free Sentinel-2 multispectral data observed closest to the sensing time of DESIS spectra were also downloaded. These Sentinel spectra were downloaded as Level-2A surface reflectance with atmospheric effects being corrected. Each spectrum consisted of 12 bands covering the visible and near-infrared spectral regions. The Sentinel-2 data were re-sampled into a spatial resolution of 30 m to be consistent with that of the DESIS data. A comparison between the DESIS and Sentinel-2 spectra at one of the ground sampling plots is shown in Fig. \ref{fig:compare}. Both spectra cover the visible, near-infrared, and short-wave-infrared (SWIR) regions, including the red-edge region that is critical for vegetation mapping. The Sentinel-2 and DESIS spectra show a consistent shape in these spectra regions, with DESIS having a much denser band coverage. The SWIR region is covered by Sentinel-2 only with its two bands. Though information in SWIR is not contained in the DESIS data, it will be provided by the upcoming DLR mission of EnMAP (for which DESIS has been served as a preparation mission), as shown in Fig. \ref{fig:compare}.

\subsection{Dimensionality Reduction for DESIS Hyperspectral Data}
\label{ssec:dimensionality_reduction}

The DESIS hyperspectral data provide rich information in the abundant and spectrally continuous bands, but these bands are highly collinear. We performed dimensionality reduction to address this problem---testing three different approaches, namely the Principal Component Analysis (PCA), Canonical Correlation Analysis (CCA), and Partial Least Squares analysis (PLS). These methods aim to find a linear transformation to project the DESIS spectra from the original space of $n$ spectral bands to a new space of a reduced dimensionality defined by $k$ uncorrelated components, with $k$ being smaller than $n$. 

Mathematically, the transformation for dimensionality reduction is written as:
\begin{equation}\label{eq:transformation}
\mathbf{T}=\mathbf{X}\mathbf{W},
\end{equation}
where the input data for the transformation is $\mathbf{X}$, which is a $n\times m$ matrix consisting of the original spectra, with $n$ being the number of observed spectra and $m$ being the number of spectral bands; the output of the transformation is $\mathbf{T}=[{\bm{t}}_1,{\bm{t}}_2, \cdots, {\bm{t}}_k]$, which is a $n\times k$ matrix consisting of $k$ components of $\mathbf{X}$; $\mathbf{W}=[{\bm{w}}_1,{\bm{w}}_2, \cdots, {\bm{w}}_k]$ is a $m\times k$ matrix transforming $\mathbf{X}$ from the original $n$-dimensional space into a new space of $k$ components. The weights in $\bm{w}_i$ are a measure of relative contributions of the original bands to the transformed component ${\bm{t}}_i=\mathbf{X}{\bm{w}}_i$. The number of components, $k$, needs to be preset. In this study, $k$ is selected as the one that achieves the highest cross-validation accuracy.

The PCA is an unsupervised algorithm that finds orthogonal components as the ones that explain the maximum variance in the spectral data, disregarding the target values of species richness. In contrast, CCA and PLS are supervised with both spectral data and species richness values being taken into account in the computation of components. The difference between CCA and PLS is that CCA seeks to maximise the correlation between computed components and species richness values, while PLS aims to maximise the covariance between the two.

\subsection{Estimation of Species Richness with Regression Models}

After reducing the dimensionality of spectral data from the original $n$ bands to $k$ components, regression is conducted to predict species richness from the components, such that the mismatch between model-predicted and ground-truth species richness is minimised. The Kernel Ridge Regression (KRR), Gaussian Process Regression (GPR), and Random Forest Regression (RFR) algorithms are employed and compared, covering respectively the deterministic, Bayesian, and ensemble approaches to statistical regression. Here we focus on formulating our task of estimating species richness within the frameworks of these regression approaches.

The DESIS spectra $\left \{ \bm{x}_i \right \}_{i=1}^n$ can be represented by their components $\left \{ \bm{t}_i \right \}_{i=1}^n$ using one of the dimensionality reduction methods described in Subsection \ref{ssec:dimensionality_reduction}. The KRR transforms $\left \{ \bm{t}_i \right \}_{i=1}^n$ into a feature space of high dimensionality (potentially infinite dimensionality) via a function $\varphi(\bm{t}_i)$. A linear model in the high-dimensional feature space is non-linear when it projects back to the original space, thus enabling capturing non-linear relationships in the data. The number of parameters for a linear model increases with space dimensionality, posing a risk of over-fitting. In order to constrain the model complexity, a regularisation term is adopted in KRR to penalise the norm of the coefficient vector $\bm{b}$. The optimisation problem is:
\begin{equation}\label{eq:krr}
\arg\min_{\bm{b}} \quad \sum_{i=1}^{n} (\bm{t}_i^{\text{T}}\bm{b}-y_i)^2 + \lambda \left \| \bm{b}  \right \|_{2} ^ {2},
\end{equation}
where $\lambda$ is the regularisation parameter controls the relative importance of the regularisation term $\left \| \bm{b}  \right \|_{2} ^ {2}$. 

The input data for Eq. \ref{eq:krr} are the transformed components $\left \{ \bm{t}_i \right \}_{i=1}^n$. Solving Eq. \ref{eq:krr} does not involve calculating $\varphi(\bm{t}_i)$. Instead, it only requires computation of the inner product $k(\bm{t}_i, \bm{t}_j) = \varphi(\bm{t}_i)\varphi(\bm{t}_j)$, where $\bm{t}_i$ and $\bm{t}_j$ are pairs from the input data $\left \{ \bm{t}_i \right \}_{i=1}^n$. In this study, the kernel function $k(\bm{t}_i, \bm{t}_j)$ is specified as a combination of a dot-product kernel $k_{d}(\bm{t}_i, \bm{t}_j)$, a radial-basis function kernel $k_{r}(\bm{t}_i, \bm{t}_j)$, and a white kernel $k_{w}(\bm{t}_i, \bm{t}_j)$:
\begin{equation}
\label{eq:kernel_function}
\begin{split}
k(\bm{t}_i, \bm{t}_j) & = k_{d}(\bm{t}_i, \bm{t}_j) + k_{r}(\bm{t}_i, \bm{t}_j) + k_{w}(\bm{t}_i, \bm{t}_j), \\
k_{d}(\bm{t}_i, \bm{t}_j) & = \bm{t}_i \cdot \bm{t}_j + \sigma^2, \\
k_{r}(\bm{t}_i, \bm{t}_j) &= {\rm exp}(-\frac{\left \| \bm{t}_i - \bm{t}_j \right \|_2^2}{2l^2}), \\
k_{w}(\bm{t}_i, \bm{t}_j) &= \delta \quad {\rm if} \; \; \bm{t}_i = \bm{t}_j \; {\rm else} \; \; 0,
\end{split}
\end{equation}
where $\sigma$, $l$, and $\delta$ are hyperparameters that need to be selected with grid search. The dot-product and radial-basis function kernels account for the linearity and non-linearity of the data, respectively, while the white kernel explains the noise in the data. For a new spectrum $\bm{x}$ with its extracted components $\bm{t}$, the predicted species richness is $\hat{y}=\sum_{i=1}^{n}{\alpha}_{i}k(\bm{t}_i, \bm{t})$, where ${\alpha}_{i}=(\mathbf{K}+\lambda \mathbf{I})^{-1}y_i$ with $\mathbf{K}_{i,j} = k(\bm{t}_i, \bm{t}_j)$.

In contrast to KRR that is formulated in a deterministic form, GPR is a Bayesian approach for regression. The underlying function correlating DESIS components and species richness is assumed to be distributed probabilistically as a Gaussian Process (GP):
\begin{equation}\label{eq:gp}
f(\bm{t})\sim \mathcal{GP}(m(\bm{t}), k(\bm{t}, \bm{t}'))
\end{equation}
where $m(\bm{t})$ is the mean function which is often set to 0, and $k(\bm{t}, \bm{t}')$ is the covariance function that can be specified as a kernel function. The input data for Eq. \ref{eq:gp} are the transformed components $\left \{ \bm{t}_i \right \}_{i=1}^n$. The covariance function $k(\bm{t}, \bm{t}')$ for GPR is set in the same way as that for KRR (Eq. \ref{eq:kernel_function}), i.e., $k(\bm{t}, \bm{t}')=k(\bm{t}_i, \bm{t}_j)$ where $\bm{t}_i$ and $\bm{t}_j$ are pairs from the input data $\left \{ \bm{t}_i \right \}_{i=1}^n$. In contrast to KRR in which the optimal kernel parameters is found by grid search, the parameters of the covariance function in GPR can be automatically determined based on gradient-ascent on the marginal likelihood function. After the posterior likelihood is determined for the GP, the predicted species richness is $\hat{y}=\sum_{i=1}^{n}{\alpha}_{i}k(\bm{t}_i, \bm{t})$ for a new spectrum, where ${\alpha}_{i}=(\mathbf{K}+\epsilon ^2 \mathbf{I})^{-1}y_i$ with $\mathbf{K}_{i,j}$ being $k(\bm{t}_i, \bm{t}_j)$ and $\epsilon$ being a pre-set parameter explaining the noise in the data.

In addition to KRR and GPR that are  respectively belong to the deterministic and Bayesian regression categories, we also tested the ensemble method of RFR. A random forest combines a number of decision tree regressors and takes the average regression result over all trees as the final estimate. Due to the ensemble structure, RFR tends to produce robust regression results with high resistance to overfitting and data noise. The training of a random forest regressor minimises the following optimization function:
\begin{equation}\label{eq:RFR}
\arg\min_{\bm{\theta}_{j}} \quad  \dfrac{1}{d}\sum_{j=1}^{d} \sum_{i=1}^{n} \left[h(\bm{t}_i;\bm{\theta}_{j})-y_i \right] ^2,
\end{equation}
where $h(\cdot)$ is the decision tree regressor with $\bm{\theta}_{j}$ being the parameters of the $j$th tree; $d$ is the number of decision trees, which was set to 100 in this study.

\subsection{Determination of Hyperparameters}

Hyperparameters that need to be pre-set included the number of components $k$ in the dimensionality reduction methods PCA, CCA, and PLS, and the kernel parameters $\sigma$, $l$, and $\delta$ in the non-linear regression method KRR. The number of components $k$ determines how much information to retain after dimensionality reduction. An optimal selection of $k$ is able to reduce data redundancy without excessively discarding useful information in the original DESIS spectra. In this study, we tested different $k$ values ranging from 1 to 10. The optimal $k$ value was selected based on the accuracy of species richness prediction, and the amount of variance in the spectral data that the retained components could explain. The tunable hyperparameters $\sigma$, $l$, and $\delta$ for the kernel function define the non-linearity structure of the regression model KRR. These kernel parameters were selected based on grid search. A grid of values was tested with each parameter varying from $10^{-5}$ to $10^{5}$ on a logarithmic scale. The combination of $\sigma$, $l$, and $\delta$ that produced the best performance in species richness prediction was selected.

\subsection{\label{ssec:cross_validation}Accuracy Assessment and Analyses}

A two-fold validation scheme was used in this study for assessing the modelling accuracy. For each of the Southern Tablelands and Snowy Mountains regions, the whole data set was randomly partitioned into two subsets (Subsets I and II), with Subset I dedicated to training and Subset II for validation (Round I), followed by Subset II for training and Subset I for validation (Round II). This procedure was repeated 100 times with the data set being partitioned differently each time. Correlation diagrams were plotted between the ground-truth species richness and the predicted values from the DESIS spectra. The coefficient of correlation ($r$) and Root-Mean-Square Error (RMSE) were calculated to evaluate the performance of the models. Results with different feature extraction procedures (PCA, CCA, and PLS) and regression models (KRR, GPR, and RFR) were computed and compared.

\subsection{\label{ssec:band_importance_analysis}Band Importance Analysis}

The DESIS data consist of spectral measurements over the visible and near-infrared bands from 400 nm to 1000 nm. Importance analysis was conducted in order to analyse which spectral bands provide more explanatory power than others in predicting plant species richness. We used the vector length of contribution values of each band to all components used in the regression weighted by the partial correlation coefficients as the importance index of the band:
\begin{equation}\label{eq:band_importance}
I_{i}=\sqrt{\sum_{j=1}^{k}(w_{i,j}\cdot p_{j})^{2}},
\end{equation}
where $I_{i}$ is the importance index for the $i$th band; $w_{i,j}$ is the $(i,j)$th element of the weight matrix $\mathbf{W}$ in Eq. \ref{eq:transformation}, representing the contribution of the $i$th band to the $j$th component; $p_{j}$ is the partial correlation coefficient of the $j$th component with the target variable of species richness; $k$ is the number of components used in regression. In our study, the importance indices $ I_{i}$ are normalised to relative values $\tilde{I}_{i}$ with sum over all bands equal to one:
\begin{equation}\label{eq:normalised_importance}
\tilde{I}_{i} = \frac{I_{i}}{\sum_{i=1}^{m}I_{i}},
\end{equation}
where $m$ is the number of spectral bands.

\subsection{\label{ssec:comparison_with_multispectral_data}Comparison with Multispectral Data}

Spaceborne multispectral imagery such as Sentinel-2 is more readily available than hyperspectral. Though Sentinel-2 images have less bands than the DESIS hyperspectral data, they are delivered on a more stable and systematic basis. In this study, an analysis is conducted to see if Sentinel-2 multispectral data are able to achieve comparable results. Through this analysis, we hoped to examine how well Sentinel-2 could serve as a substitute for plant biodiversity mapping in instances where hyperspectral data are unavailable.

The Sentinel-2 data set described in Section \ref{ssec:satellite_data} was used for the comparison. In order to conduct a fair comparison between hyperspectral and multispectral data with minimised differences in instrumental specifications and acquisition conditions, a Sentinel-2-like synthetic data set was simulated from the DESIS data. The simulation involved resampling the DESIS spectra using the spectral response functions of Sentinel-2's visible and near-infrared bands. Both the real and synthetic Sentinel-2 sets were used for species richness prediction, with results being compared to those achieved with DESIS data.

\section{Results}

\subsection{Spectral Reflectance Differences Between Species Richness Classes}

\begin{figure}[htb!]
\centering
\captionsetup{font=normalsize}
\includegraphics[width=11cm]{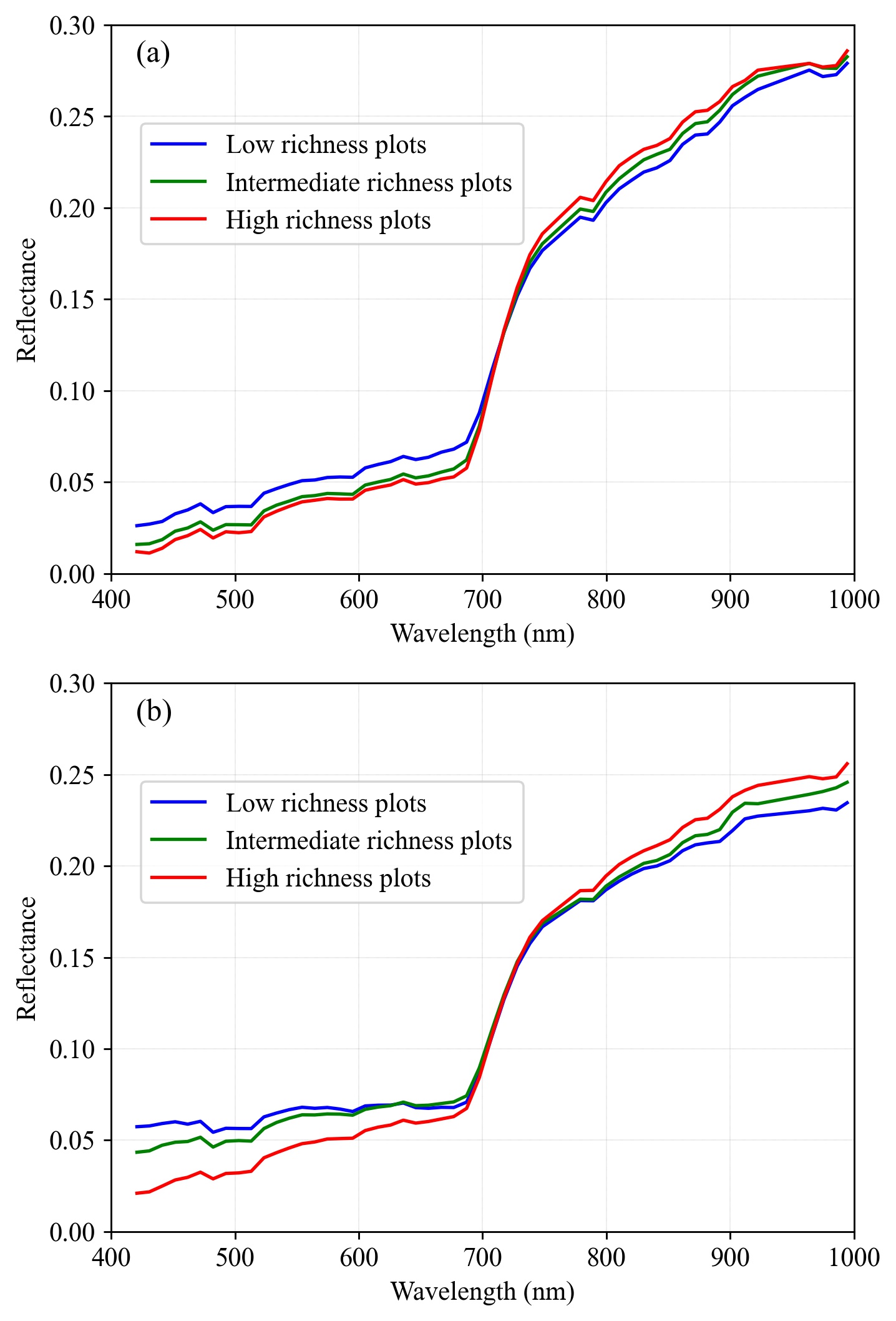}
\caption{Average DESIS reflectance spectra calculated from field sample plots with low, intermediate, and high species richness for the (a) Southern Tablelands and (b) Snowy Mountains regions. \label{fig:different_richness}}
\end{figure}

The blue, green, and red curves in Fig. \ref{fig:different_richness} show the DESIS spectra averaged over ground sampling plots with species richness falling into the low, intermediate, and high tertiles, respectively. Results for the Southern Tablelands and Snowy Mountains regions are displayed in Figs \ref{fig:different_richness}a and b, respectively. For each region, it was observed that plots of higher richness showed a lower reflectance in the visible range of $400\sim680$ nm. Considering that major absorption features of chlorophyll are located within the visible region \citep{zhao2014earlya}, the lower reflectance in this spectral portion might indicate a higher concentration of chlorophyll. It was also observed that plots of higher richness showed a higher reflectance in the near-infrared plateau of $780\sim1000$ nm and a steeper red edge between $680\sim780$ nm, which might suggest a larger Leaf Area Index (LAI) \citep{delegido2013red} and a greater vegetation vigor \citep{boochs1990shape} for those plots. On the basis of these observations, the spectral shape of high richness plots, as compared with spectra of intermediate and low richness plots, may imply a generally richer vegetation. This is consistent with findings reported in literature that high species richness enhances primary productivity \citep{wang2016seasonal, grace2016integrative} and biomass \citep{malhi2020synergetic, tilman1997influence}.

\subsection{Hyperparameter Selection Results}

\begin{figure}[htbp!]
\centering
\captionsetup{font=normalsize}
\includegraphics[width=9cm]{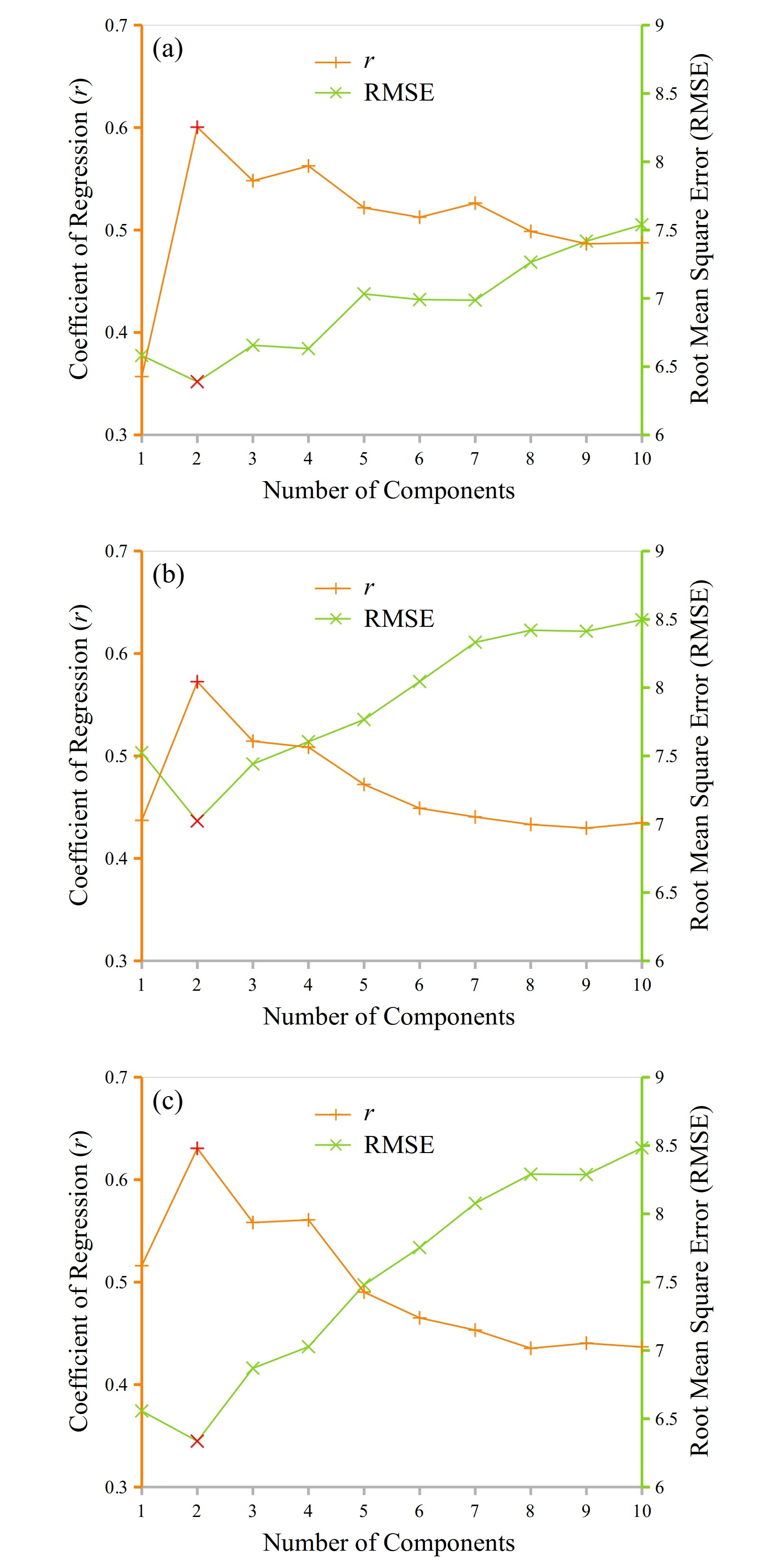}
\caption{Impact of number of components on the estimation accuracy of plant species richness with (a) Principal Component Analysis (PCA), (b) Canonical Correlation Analysis (CCA), and (c) Partial Least Squares analysis (PLS).\label{fig:components}}
\end{figure}

\begin{figure}[htb!]
\centering
\captionsetup{font=normalsize}
\includegraphics[width=13cm]{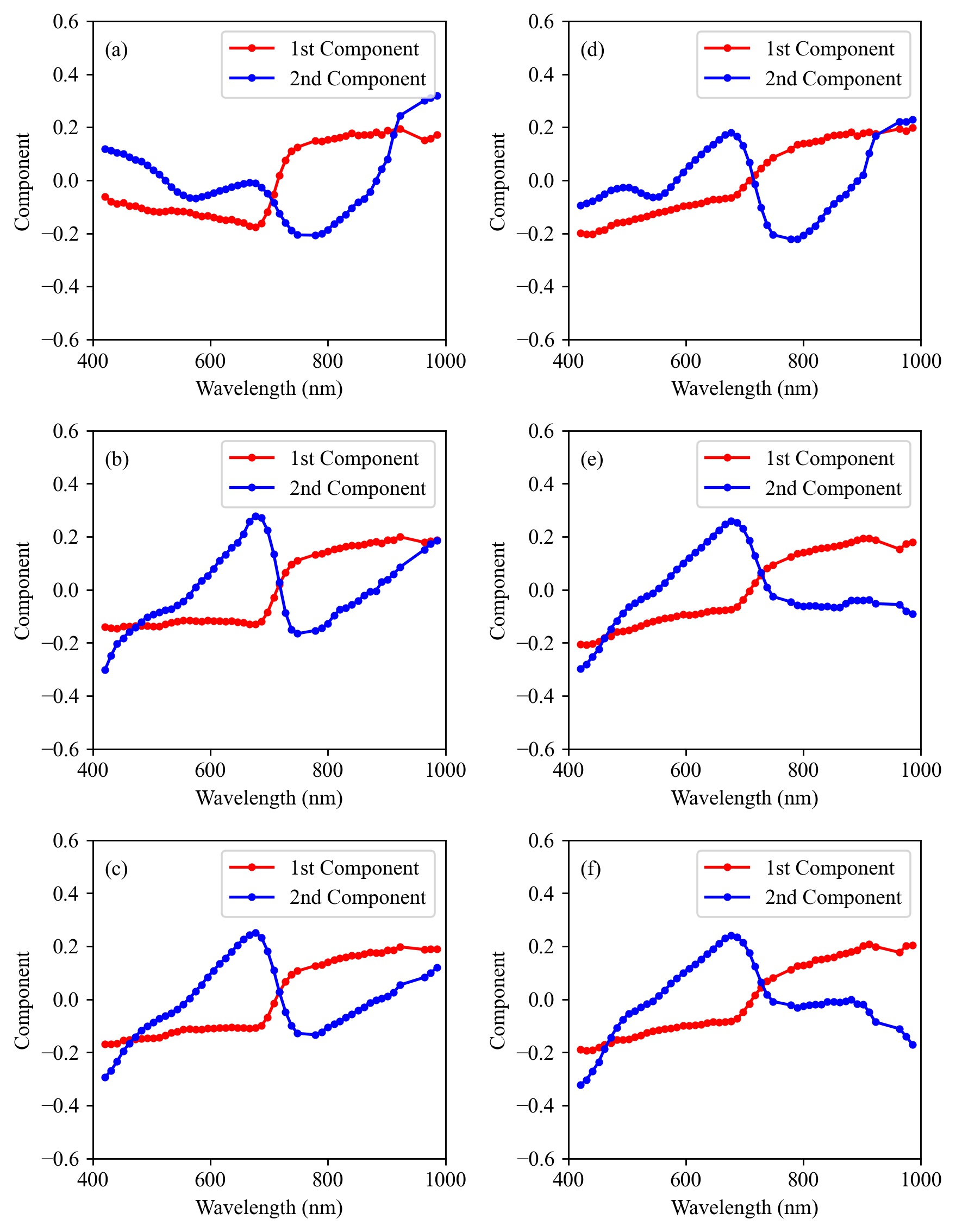}
\caption{The first and second components for the transformed DESIS spectra. Left column: Result for the Southern Tablelands with (a) Principal Component Analysis (PCA), (b) Canonical Correlation Analysis (CCA), and (c) Partial Least Squares analysis (PLS); Right column: Result for the Snowy Mountains region with (d) PCA, (e) CCA, and (f) PLS. \label{fig:component_12}}
\end{figure}

Fig. \ref{fig:components} shows the $r$ and RMSE values achieved with different numbers of components (ranging from 1 to 10) being selected as features. The results were averaged over the two study regions. It was observed that two components achieved the best performance for all the three feature extraction methods of PCA (Fig. \ref{fig:components}a), CCA (Fig. \ref{fig:components}b), and PLS (Fig. \ref{fig:components}c). When only one component was used with more information in the original spectral data being discarded, lower $r$ values and higher RMSE values were also observed, indicating a poorer performance compared with that achieved by two components. When more than two components were retained with a higher degree of data redundancy presenting, weaker results were also observed. The eigenvalue, percentage of explained variance, and cumulative percentage of explained variance with different components are shown in Table \ref{tab:eigen}. It was seen that with two components, the PCA, CCA, and PLS could explain 93.81$\%$, 89.34$\%$, and 87.38$\%$ of variance in the DESIS data, respectively. Based on these results, the number of components, $k$, was set to two in our experiments for dimensionality reduction with PCA, CCA, and PLS. The first and second components are plotted in Fig. \ref{fig:component_12}. The first component depicted the general shape of the spectral brightness, while the second component highlighted more on local spectral features of the spectrum.

\begin{landscape}

\begin{table}[htbp!]
\captionsetup{font=normalsize}
\caption{The eigenvalue, percentage of explained variance, and cumulative percentage of explained variance for components produced with Principal Component Analysis (PCA), Canonical Correlation Analysis (CCA), and Partial Least Squares analysis (PLS). \label{tab:eigen}}
\renewcommand{\arraystretch}{1.1}
\begin{tabular}{cccccccccc}
\toprule
\multirow{2}{*}{Component} & \multicolumn{3}{c}{PCA} & \multicolumn{3}{c}{CCA} & \multicolumn{3}{c}{PLS} \\
& \multicolumn{1}{c}{Eigenvalue} & \multicolumn{1}{c}{\shortstack{\% of \\ Variance}} & \multicolumn{1}{c}{\shortstack{Cumulative \% \\ of Variance}} & \multicolumn{1}{c}{Eigenvalue} & \multicolumn{1}{c}{\shortstack{\% of \\ Variance}} & \multicolumn{1}{c}{\shortstack{Cumulative \% \\ of Variance}} & \multicolumn{1}{c}{Eigenvalue} & \multicolumn{1}{c}{\shortstack{\% of \\ Variance}} & \multicolumn{1}{c}{\shortstack{Cumulative \% \\ of Variance}} \\ 
\hline
1 &44.35 &84.48 &84.48 &40.43 &74.61 &74.61 &40.96 &75.58 &75.58 \\
2 &4.90 &9.34 &93.81 &7.98 &14.73 &89.34 &6.39 &11.80 &87.38 \\
3 &2.22 &4.23 &98.04 &4.16 &7.67 &97.01 &4.94 &9.12 &96.50 \\
4 &0.44 &0.84 &98.87 &0.61 &1.13 &98.14 &0.70 &1.29 &97.79 \\
5 &0.22 &0.41 &99.29 &0.48 &0.89 &99.03 &0.41 &0.75 &98.55 \\
6 &0.10 &0.18 &99.47 &0.12 &0.23 &99.26 &0.23 &0.42 &98.97 \\
7 &0.04 &0.08 &99.55 &0.10 &0.19 &99.45 &0.14 &0.26 &99.22 \\
8 &0.04 &0.08 &99.63 &0.08 &0.14 &99.59 &0.11 &0.19 &99.42 \\
9 &0.03 &0.06 &99.69 &0.06 &0.11 &99.69 &0.08 &0.14 &99.56 \\
10 &0.02 &0.04 &99.73 &0.04 &0.08 &99.77 &0.06 &0.11 &99.67 \\ 
\bottomrule
\end{tabular}
\end{table}

\end{landscape}

\begin{figure}[htb!]
\centering
\captionsetup{font=normalsize}
\includegraphics[width=15.2cm]{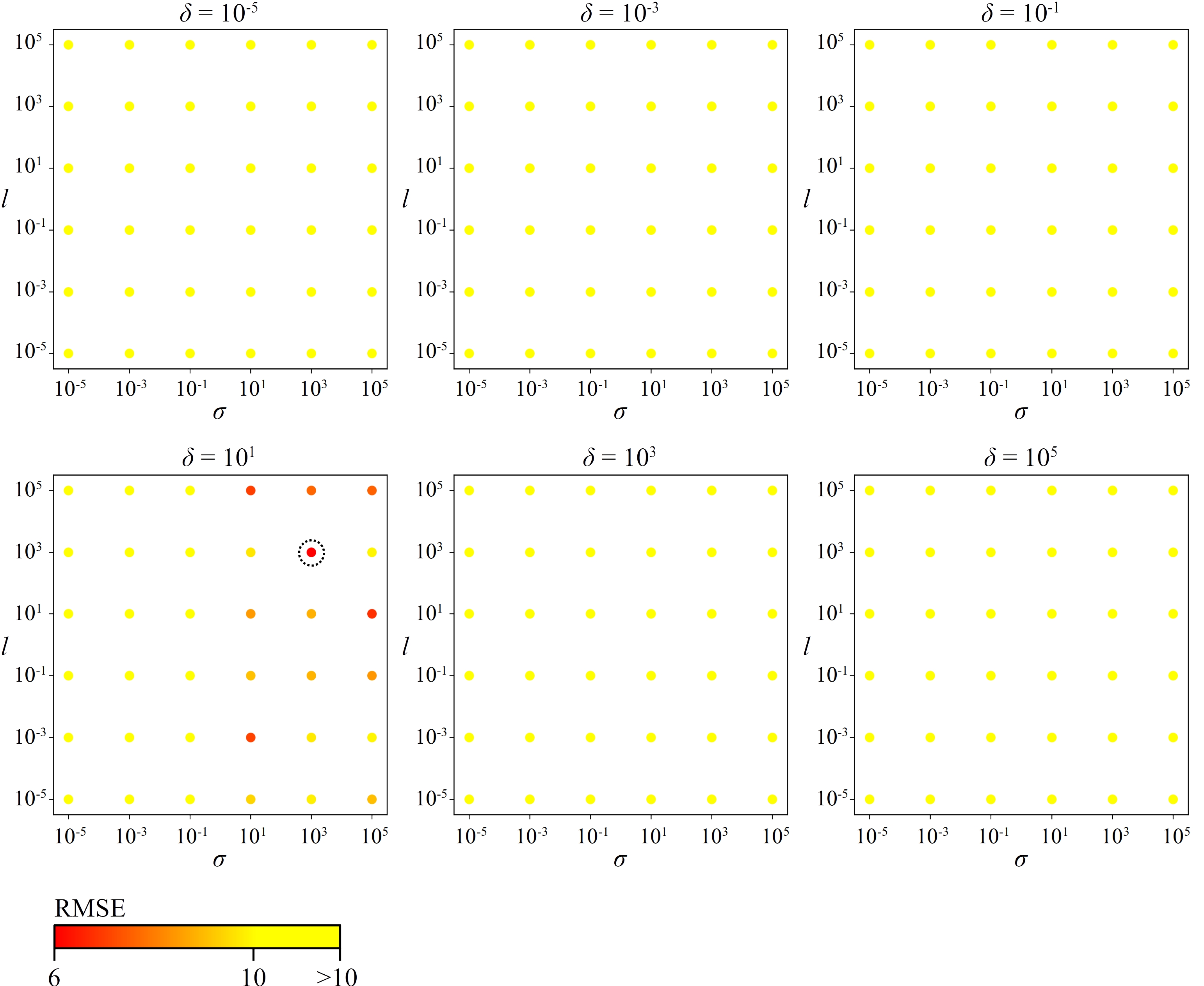}
\caption{Selection of kernel parameters $\sigma$, $l$, and $\delta$ based on grid search for Kernel Ridge Regression (KRR). The best performing combinations of kernel parameters are circled in black. \label{fig:kernel_params}}
\end{figure}

The grid searching result for the optimal KRR hyperparameters $\sigma$, $l$, and $\delta$ is shown in Fig. \ref{fig:kernel_params}. The best combination of kernel parameters was $\sigma = 10^{3}$, $l=10^{3}$, and $\delta=10$. This combination of kernel parameter values was used as default values for KRR in our experiments. 

\subsection{Plant Species Richness Prediction}

The assessment of different dimensionality reduction and regression methods (Table {\ref{tab:southern_tablelands}}) showed that, for the Southern Tablelands region, the best result was achieved with a combination of PLS for dimensionality reduction and GPR for regression, with $r$ being 0.76 and RMSE being 5.89. Results with PLS as the dimensionality reduction method performed better than those with PCA and CCA. The PLS-based models achieved $0.75\sim 0.76$ for $r$ and $5.89\sim 5.92$ for RMSE, respectively, better than the PCA-based models with $0.70\sim 0.71$ for $r$ and $5.99 \sim 6.02$ for RMSE, and the CCA-based models with $0.69\sim 0.73$ for $r$ and $5.98 \sim 6.02$ for RMSE.

\begin{table}[htb!]
\captionsetup{font=normalsize}
\caption{The coefficient of correlation ($r$) and Root-Mean-Square Error (RMSE) between model predicted and ground truth plant species richness for the Southern Tablelands region.\label{tab:southern_tablelands}}
\begin{tabularx}{\textwidth}{p{3.5cm}p{3.5cm}p{3.5cm}p{3.5cm}}
\toprule
\textbf{Dimensionality \quad Reduction}	& \textbf{Regression}	& \textbf{$r$} & \textbf{RMSE}\\
\midrule
PCA                      & KRR          & 0.70          & 6.01          \\
PCA                      & GPR          & 0.71          & 5.99          \\
PCA                      & RFR       & 0.70          & 6.02          \\
CCA                      & KRR          & 0.69          & 6.02          \\
CCA                      & GPR          & 0.72          & 5.98          \\
CCA                      & RFR       & 0.73          & 6.00          \\
PLS             & KRR & 0.75 & 5.92 \\
PLS                      & GPR          & 0.76          & 5.89          \\
PLS                     & RFR       & 0.75          & 5.91          \\
\bottomrule
\end{tabularx}
\end{table}

The assessment results for the Snowy Mountains region in shown in Table \ref{tab:snowy_mountains}. The Snowy Mountains region is located at high altitudes with less human interference and a generally lower species richness than the Southern Tablelands (Fig. \ref{fig:hist}). Compared with the results for the Southern Tablelands in Table \ref{tab:southern_tablelands}, it can be seen that generally the $r$ values were lower and RMSE were higher for the Snowy Mountains region (Table \ref{tab:snowy_mountains}). The best $r$ and RMSE were 0.68 and 5.95, respectively, achieved with a combination of PLS for dimensionality reduction and RFR for regression.

\begin{table}[htb!]
\captionsetup{font=normalsize}
\caption{The coefficient of correlation ($r$) and Root-Mean-Square Error (RMSE) between model predicted and ground truth plant species richness for the Snowy Mountains region.\label{tab:snowy_mountains}}
\begin{tabularx}{\textwidth}{p{3.5cm}p{3.5cm}p{3.5cm}p{3.5cm}}
\toprule
\textbf{Dimensionality \quad Reduction}	& \textbf{Regression}	& \textbf{$r$} & \textbf{RMSE}\\
\midrule
PCA                      & KRR        & 0.51          & 6.17          \\
PCA                      & GPR        & 0.54          & 6.03          \\
PCA                      & RFR       & 0.56          & 6.01          \\
CCA                      & KRR        & 0.52          & 6.14          \\
CCA                      & GPR        & 0.53          & 6.08          \\
CCA                      & RFR       & 0.54          & 5.99          \\
PLS                      & KRR        & 0.64          & 6.08          \\
PLS             & GPR & 0.66 & 5.97 \\
PLS                      & RFR       & 0.68          & 5.95          \\
\bottomrule
\end{tabularx}
\end{table}

\begin{figure}[htb!]
\centering
\includegraphics[width=11cm]{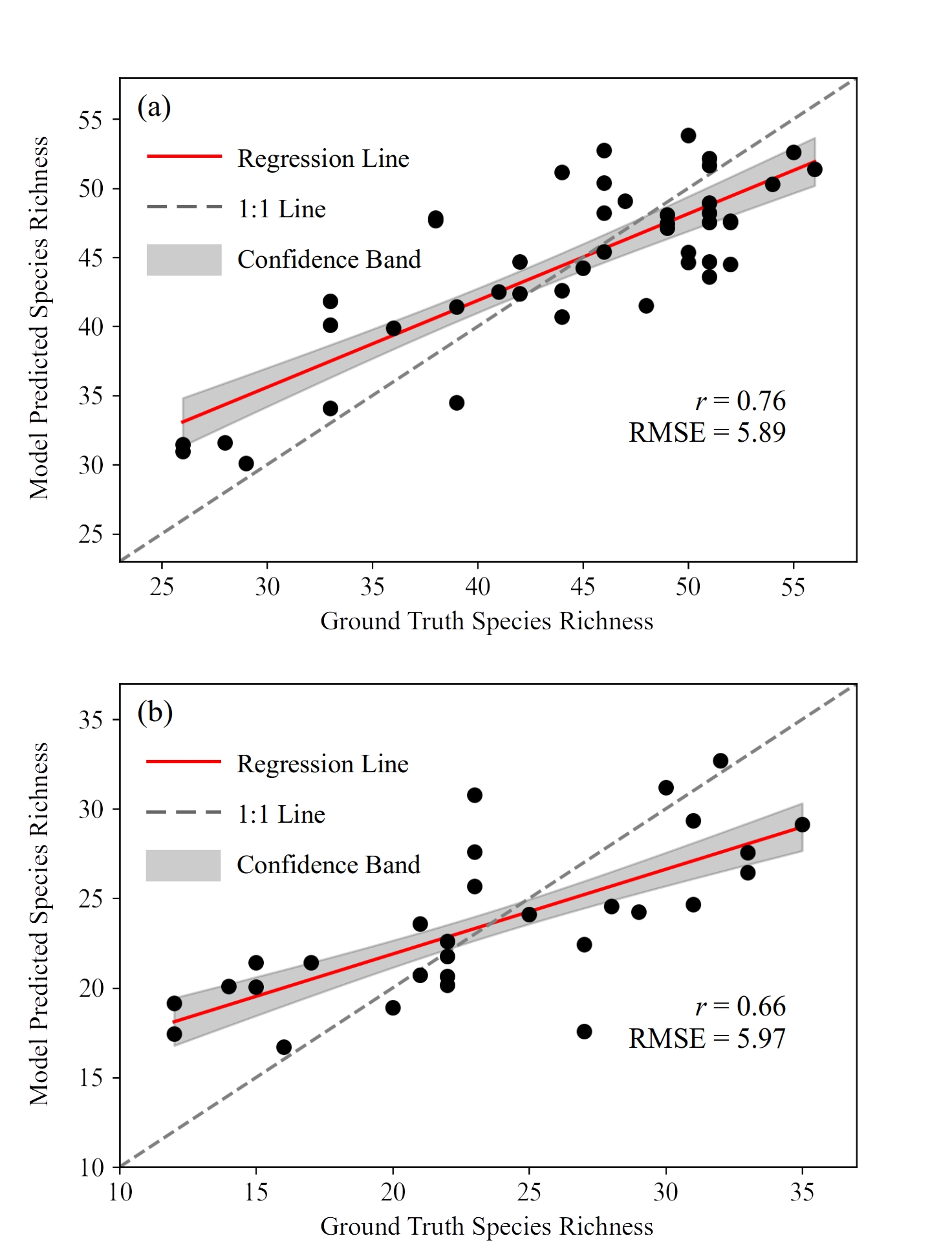}
\captionsetup{font=normalsize}
\caption{Correlation diagrams between the ground-truth species richness and the predicted values from the DESIS spectra for the (a) Southern Tablelands and (b) Snowy Mountains regions. The $r$ and RMSE stand for the coefficient of correlation and root-mean-square error. \label{fig:hyper}}
\end{figure}

The correlation diagrams in Fig. \ref{fig:hyper} show the relationship between the ground-truth species richness and the predicted values from the DESIS spectra, with data samples from the validation set. These results were produced with the best performing models in Tables \ref{tab:southern_tablelands} and \ref{tab:snowy_mountains}. 

\subsection{Generalised Modelling Results}

\begin{figure}[htb!]
\centering
\includegraphics[width=11cm]{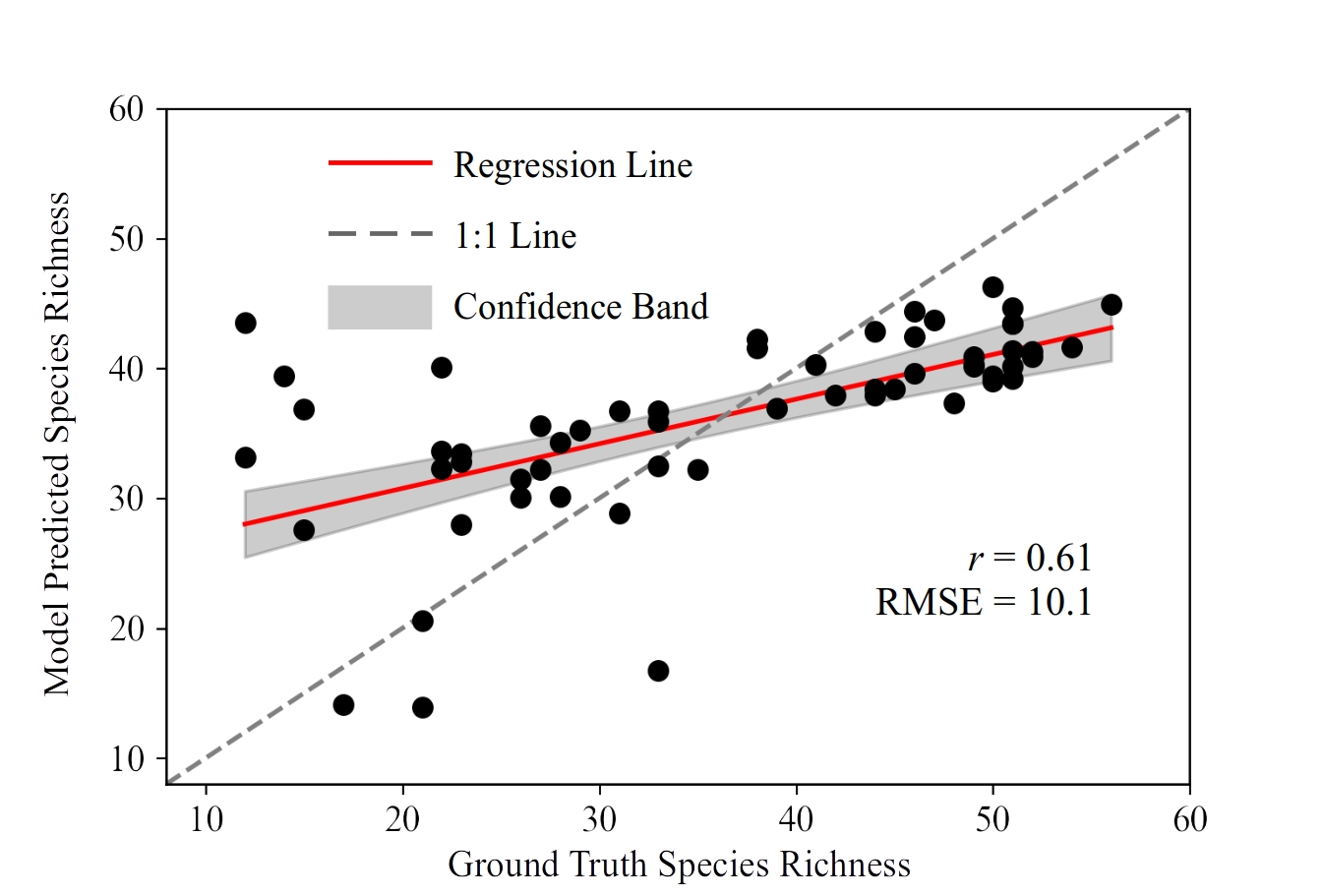}
\captionsetup{font=normalsize}
\caption{Correlation diagrams between the ground-truth species richness and values predicted from DESIS spectra with data congregating both of the Southern Tablelands and Snowy Mountains regions. The $r$ and RMSE stand for the coefficient of correlation and root-mean-square error.\label{fig:both}}
\end{figure}

Figs \ref{fig:hyper}a and \ref{fig:hyper}b show results with modelling being conducted for the Southern Tablelands and Snowy Mountains regions separately. When we modelled the pooled data from both regions, we found that accuracy decreased (Fig. \ref{fig:both}). The prediction result was $r=0.61$ and $\text{RMSE}=10.1$, which was lower than modelling with data from only one region (Figs \ref{fig:hyper}a and \ref{fig:hyper}b). This indicates that location-specific modelling performs better than using one model to describe multiple regions. 

The difficulty of modelling relationships between hyperspectral data and plant species richness for multiple regions, compared modelling for each region separately, is worth further investigation. Different regions may differ in the assemblages of plant species, and their compositional and structural properties, resulting in extra variations in hyperspectral data in addition to those induced by the richness of species. These additional variations may add complexity in exploring useful information in hyperspectral data to predict plant species richness. The location-specific relationship hyperspectral data and plant species richness calls for location-dependant modelling or encoding location information into the input spectra in future studies when mapping plant species richness at continental or global scales is attempted. 

\subsection{The Relative Importance of Spectral Bands}\label{ssec:relative}

\begin{figure}[htb!]
\centering
\includegraphics[width=13.5cm]{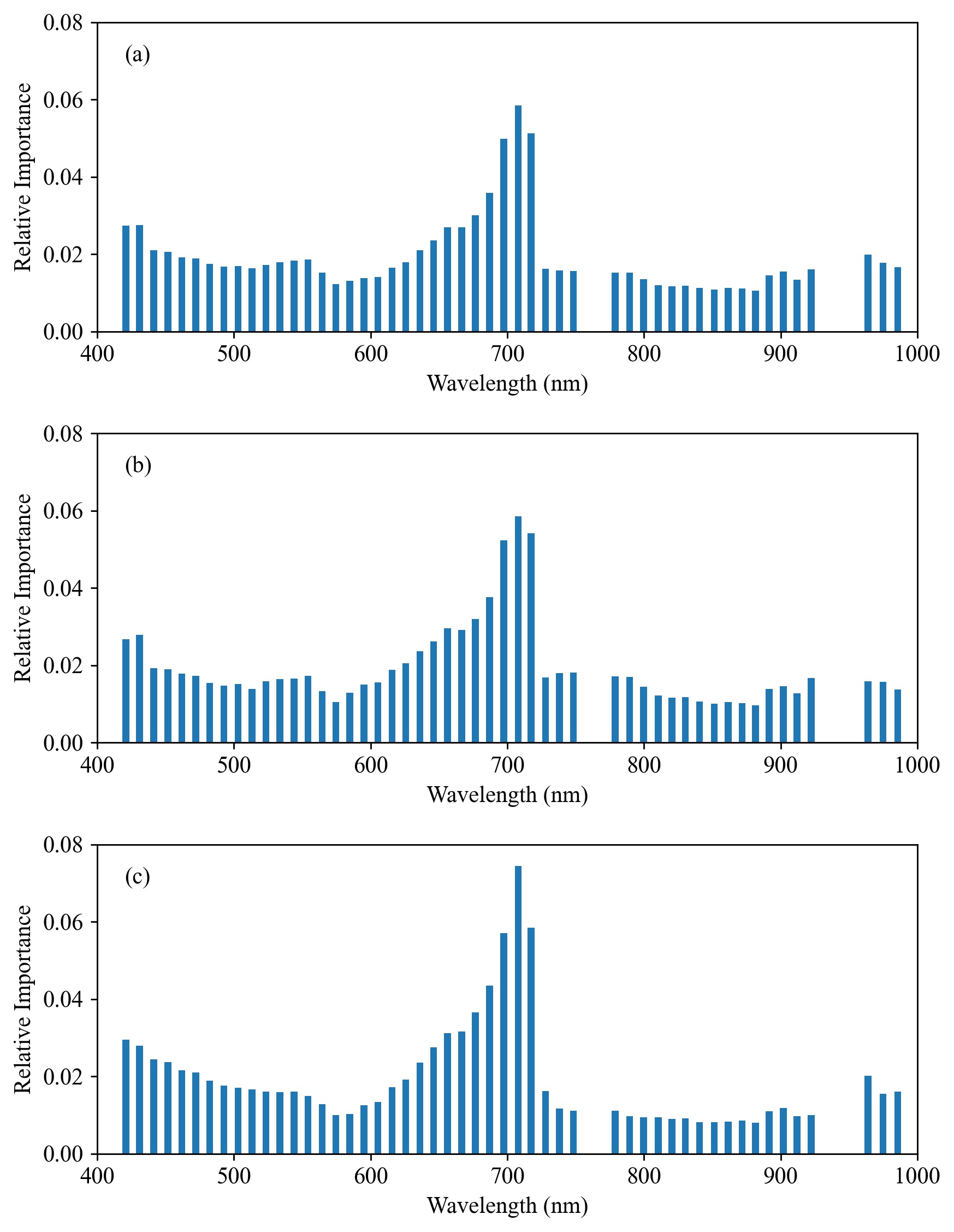}
\captionsetup{font=normalsize}
\caption{Relative importance analysis for the DESIS bands in predicting plant species richness with spectral features being extracted by (a) Principal Component Analysis (PCA), (b) Canonical Correlation Analysis (CCA), and (c) Partial Least Squares analysis (PLS). \label{fig:relative}}
\end{figure}

The relative importance of DESIS bands in predicting plant species richness is shown in Fig. \ref{fig:relative}, with the aim to analyse which parts of the spectrum had the most explanatory power. Subplots \ref{fig:relative}a, b, and c display the results with PCA, CCA, and PLS being used as the feature extraction procedure, respectively. From these subplots, it was observed that bands in the red-edge spectral region of approx. 700 $\sim$ 720 nm showed the highest importance. This may suggest that the slope of the red-edge is an important spectral feature for plant species richness prediction, considering that the low, intermediate, and high species richness plots showed varied red-edge slopes in Fig. {\ref{fig:different_richness}}. This observation might be supported by literature that the red-edge is a critical spectral region for vegetation mapping as it is closely related to biological variables such as leaf area index \citep{delegido2013red}, plant vigour \citep{boochs1990shape}, and biochemical contents \citep{mutanga2007red}. Followed by the red-edge, the visible range of approx. 400 $\sim$ 700 nm also showed high importance. The importance of visible bands in plant species richness prediction might be justified by the fact that, this portion of the spectrum, especially bands in red and blue, is the major leaf pigment absorption range. It is sensitive to mainly chlorophyll a and b contents, according to the sensitivity analysis result reported in \cite{zhao2014earlya, zhao2014earlyb}. Though less important than the red-edge and visible regions, the near-infrared region with wavelengths longer than 720 nm also showed some explanatory power. This indicated that the near-infrared region also provided contributory information in predicting plant species richness, as this portion of the spectrum, often characterised by high reflectance for vegetation, is sensitive to leaf thickness \citep{zhao2014earlyb} and the amount, arrangement, and inclination of leaves in the canopy \citep{knipling1970physical}.

\subsection{Comparison with Multispectral Data}

\begin{figure}[htb!]
\centering
\includegraphics[width=11cm]{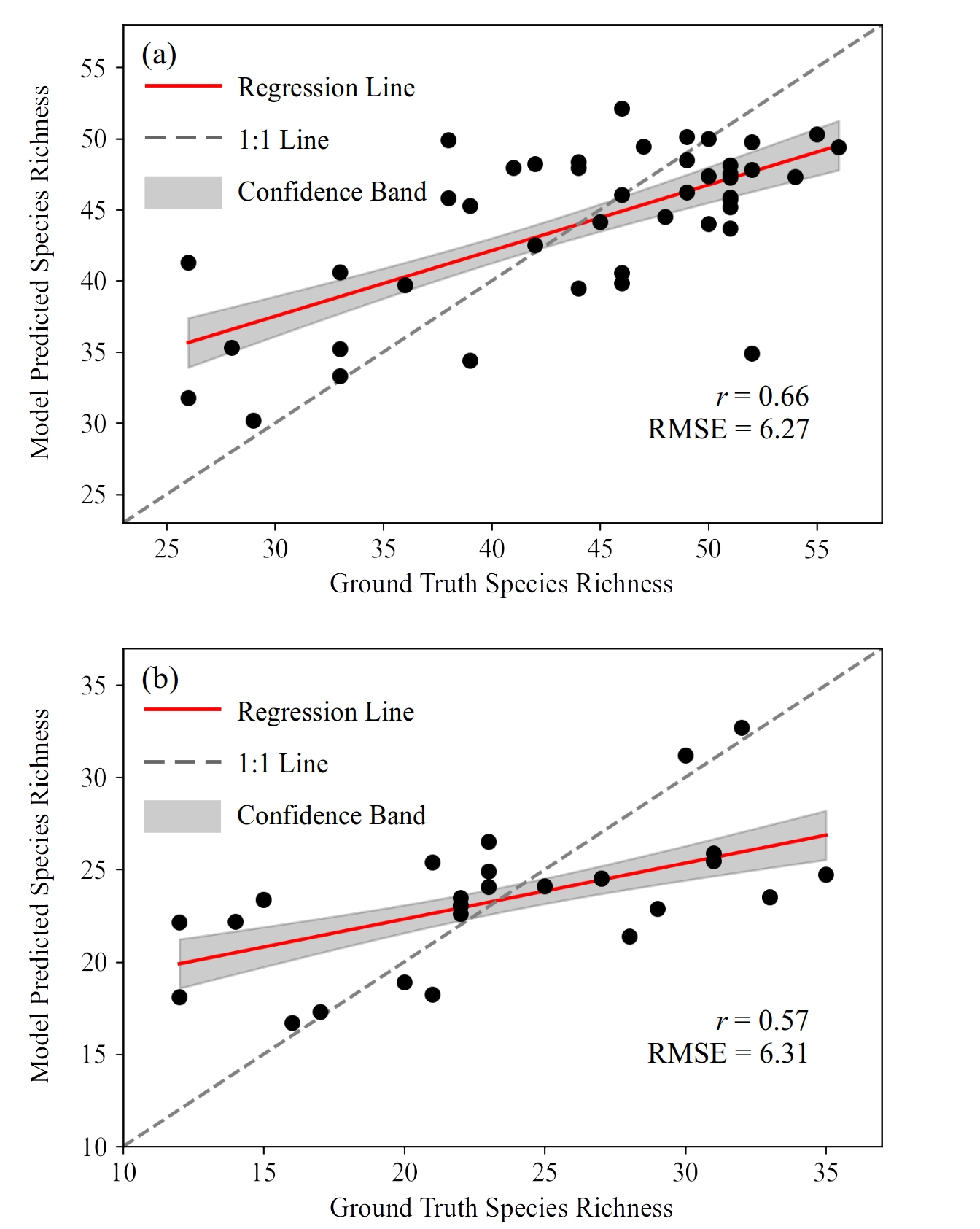}
\captionsetup{font=normalsize}
\caption{Correlation diagrams between the ground-truth species richness and values predicted from the real Sentinel-2 multispectral data set for the (a) Southern Tablelands and (b) Snowy Mountains regions. The $r$ and RMSE stand for the coefficient of correlation and root-mean-square error.\label{fig:multi}}
\end{figure}

\begin{figure}[htb!]
\centering
\includegraphics[width=11cm]{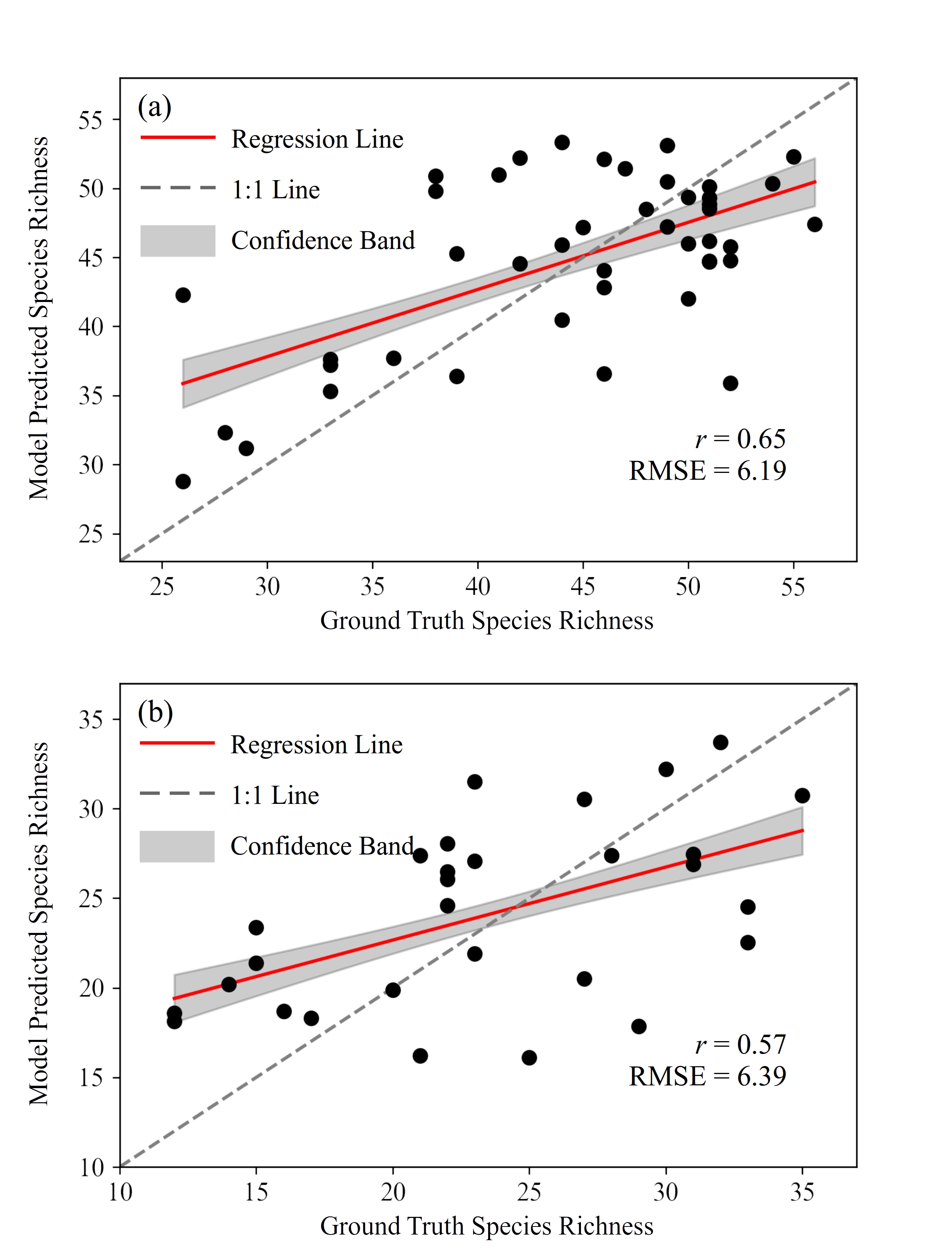}
\captionsetup{font=normalsize}
\caption{Correlation diagrams between the ground-truth species richness and values predicted from the Sentinel-2-like synthetic multispectral data set for the (a) Southern Tablelands and (b) Snowy Mountains regions. The $r$ and RMSE stand for the coefficient of correlation and root-mean-square error. \label{fig:synthetic}}
\end{figure}

Figs \ref{fig:multi}a and \ref{fig:multi}b show correlation diagrams between the ground-truth species richness and values predicted from the real Sentinel-2 multispectral data for the Southern Tablelands and Snowy Mountains regions, respectively. It is seen that, a prediction result of $r=0.66$ and $\text{RMSE}=6.27$ was achieved for the Southern Tablelands region (Fig. \ref{fig:multi}a), and $r=0.57$ and $\text{RMSE}=6.31$ for the Snowy Mountains region (Fig. \ref{fig:multi}b). The prediction results with the Sentinel-2-like synthetic data set are shown in Fig. \ref{fig:synthetic}. For the Southern Tablelands, the $r$ and RMSE values are 0.65 and 6.19, while for the Snowy Mountains, the $r$ and RMSE values are 0.57 and 6.39.   

These multispectral results were slightly worse than the hyperspectral results shown in Fig. \ref{fig:hyper}. This could be explained by the fact that hyperspectral data contain richer information in the spectral domain. However, when taking data availability and stability into consideration, spaceborne multispectral imagery could serve as a reliable alternative for plant biodiversity mapping. Also, data such as Sentinel has much more consistent and regular sampling over time, and harnessing the added information in temporal changes in multispectral signal could provide further explanatory power for biodiversity analyses.

\section{Discussions}

\subsection{Comparison to Previous Studies}

Predicting plant biodiversity from remotely sensed measurements helps with making evidence-informed environmental policies and conducting effective conservation activities. Hyperspectral imagery delivered by the recently launched DESIS instrument opens the potential for plant biodiversity monitoring at a finer spatial scale and with a higher accuracy. In this study, we take the the Southern Tablelands and Snowy Mountains regions in southeast Australia as experimental sites to test the relationship between DESIS data and plant species richness. A two step approach is proposed, where feature extraction techniques are first used to reduce the dimensionality of hyperspectral data, followed by regression models with kernel functions to account for the linearity, non-linearity, and noise in data.

Obtaining informative features from hyperspectral data is important to the success of subsequent interference of plant species richness. In previous studies, features have been primarily selected as a subset of the original bands, or as spectral indices computed from a subset of the original bands (e.g., \cite{malhi2020synergetic}, \cite{peng2018assessment}, and \cite{wang2016seasonal}). The bands or indices are often selected based on our \emph{a priori} knowledge of the spectral properties, such as absorption features of biochemical contents. Though selected features usually offer good explainability, it means we have to discard information in unselected bands. Considering the large number of bands in hyperspectral data, many of them would be discarded as the high collinearity of hyperspectral data often requires a considerable reduction of dimensionality. In contrast, the feature extraction approach, as adopted in this work, makes use of the information in all original bands by transforming them into a new feature space of lower and non-collinear dimensionality.

Linear models have been primarily employed in previous studies to relate features to species richness. For example, multiple linear regression has been adopted by \cite{wang2016seasonal} and \cite{malhi2020synergetic}, and stepwise linear regression by \cite{peng2018assessment}. However, the relationship between features and richness might not necessarily  follows simply a linear pattern. In our study, a novel kernel function is proposed (Eq. \ref{eq:kernel_function}), with the dot-product and radial-basis function kernels account for the linearity and non-linearity of the data, respectively, and the white kernel explains the noise in the data. The ability of our model to explore both linear and non-linear patterns distinguishes our work from aforementioned studies.

\subsection{Mechanism of Plant Species Richness Prediction}

Though detailed field surveys have been deemed as the most accurate and reliable way for assessing plant biodiversity, remote sensing data can serve as a proxy for large-scale and cost-effective biodiversity mapping. The mechanism justifying the use of remote sensing has been widely discussed in literature. It is worth noting the exact mechanism is dependant on which method is used and the sensor specifications (e.g., pixel size and spectral range). The richness of species can be estimated either directly from raw spectra, or via the extracted plant functional traits or types \citep{wang2019remote}. Methods based on the spectral variation hypothesis is also intriguing, whereby the spectral variation across spatially adjacent pixels is employed as the proxy \citep{palmer2002quantitative, fassnachtlink}. Here we focus on discussing the underlying mechanism that underpins our study where the DESIS spectra with a pixel size of 30 m and a spectral range of 400--1000 nm are linked to on-ground species richness via feature extraction and statistical regression.

It has been reported in literature that plant communities of high diversity tend to have an enhanced primary productivity \citep{wang2016seasonal, grace2016integrative} and a higher above ground biomass \citep{malhi2020synergetic, tilman1997influence}. Though the reason to explain the richness--productivity/biomass relationship is a subject of debate, a common theory is that the complementary roles played by different species lead to lower nutrient losses and more sustainable soils \citep{tilman1996productivity}. Species complementarity allows plants to capture resources in ways that are complementary in space or/and time, leading to increased biomass production \citep{cardinale2007impacts}. For example, a high number of species allows plants to reside in various partitions of niches, resulting in a denser occupation of space and a higher efficiency of water, nutrition, and sunlight usage.

Based on this relationship, many studies have successfully estimated plant species richness from hyperspectral measurements (e.g., \cite{wang2016seasonal}, \cite{malhi2020synergetic}, and \cite{peng2018assessment}), given that remotely sensed hyperspectral data is a good proxy of vegetative biomass and primary productivity. A positive and dynamic productivity–diversity relationship is observed in a prairie grassland experiment at Cedar Creek, Minnesota, USA, with NDVI being employed as a proxy of vegetation productivity to estimate species richness \citep{wang2016seasonal}. The correlation between spectral indices and plant species diversity is also reported in \cite{peng2018assessment} for a semi-arid sandland ecosystem in Inner Mongolia, China. 

It is worth noting that previous studies are primarily focused on ground-measured (e.g., \citep{peng2018assessment} and \cite{wang2016seasonal}) or airborne (e.g., \cite{asner2008hyperspectral}) hyperspectral data, with a limited spatial range. The recently launched DESIS and PRISMA \citep{pignatti2013prisma, verrelst2021mapping}, and the upcoming EnMAP (Environmental Mapping and Analysis Program) \citep{guanter2015enmap} missions, enable us to test the potential of spaceborne hyperspectral measurements in plant species richness mapping. Our study shows a promising correlation between the two, and finds that the correlation is location-dependent.

\subsection{Limitations}

Over the past decades, in-situ samples of plant species richness have been collected via various survey campaigns. In total, more than 188 thousands of samples have been gathered in Australia as of the year of 2018 \citep{gellie2018overview}. However, most of the samples are not able to be matched with a DESIS observation that is temporally close enough to them, as DESIS has not been in operation until 2018. In this study, in order to avoid large temporal discrepancies, we have limited our on-ground samples to those that are less than three years apart with their associated DESIS spectra. The limited spatial coverage of DESIS images, and bushfires during the 2019--2020 summer, have further reduced the number of available samples. As a result, analyses in this study are conducted on a relatively small number of samples. Nevertheless, the ever increasing amount of both satellite images and on-ground samples will enable more comprehensive assessment of the potential of spaceborne hyperspectral remote sensing for plant biodiversity mapping. Moreover, though the data set used in this work does not have information on which strata the observed species come from, it is worth splitting out richness for different strata in future field surveys. With strata information on record, we might be able to model tree richness and understory richness separately, and then combine them once predicted to get total richness. The results reported in this study may serve as a basis for future studies.

It is important to note that the DESIS pixel size ($30\times30$ m) does not match exactly with the plot area of ground species richness samples ($400\;\text{m}^{2}=20\times20\;\text{m}$). Though richness measurements could be scaled to a larger or smaller plot area using an assumed power relationship ($S=cA^{z}$) between species richness ($S$) and plot area ($A$) \citep{rosenzweig1995species}, the location-specific power parameter ($z$) is often hard to be accurately determined without adequate knowledge about the experiment site. It is suggested that, in future field campaigns, it is worthwhile to gather information on the richness-area relationship, in order to facilitate accurate up- and down-scaling of plot areas. Potential inaccuracies in geo-registration of DESIS images and in geo-positioning measurements during field surveys may also result in geo-mismatch between pixels and ground plots. Better geo-registration and geo-positioning accuracies would help reduce uncertainties in predicting species richness in future works.

The DESIS sensor covers the visible and near-infrared (VNIR) portions of the spectrum. As shown in this study, this spectral range is informative in plant biodiversity mapping. However, it is also worthwhile to leverage the potential of the short-wave-infrared (SWIR) bands. The upcoming EnMAP imaging spectroscopy mission \citep{guanter2015enmap} is scheduled to be launched in 2022. The EnMAP dual-spectrometer instrument, covering both VNIR and SWIR from 420 nm to 2450 nm (as shown in Fig. \ref{fig:compare}), will provide an opportunity to integrate information from SWIR for plant biodiversity mapping. In addition to hyperspectral optical imagery, the combination of data from other sensor types, such synthetic-aperture radar (SAR) or LiDAR data, could also be explored to improve our ability in remote mapping of plant biodiversity.

\section{Conclusion}

Spaceborne hyperspectral remote sensing is a promising and cost-effective data source to enable plant biodiversity mapping. Thanks to its advanced spectral and spatial specifications, the recently launched hyperspectral instrument DESIS (the DLR Earth Sensing Imaging Spectrometer) opens up an opportunity to monitor plant biodiversity at a finer spatial scale and with a higher accuracy. In this study, we assessed the ability of DESIS hyperspectral data in predicting plant species richness in the Southern Tablelands and Snowy Mountains regions in southeast New South Wales, Australia. The spectral features were firstly extracted, and then correlated to plant species richness via statistical regression. We evaluated the performance of several combinations of feature extraction procedures (PCA, CCA, and PLS) and regression models (KRR, GPR, and RFR). The main findings of this study are summarised as follows:

(1) Plant species richness values were predicted from DESIS data in the two study regions. Prediction accuracies fell within a comparable range for different combinations of feature extraction techniques and regression models (Tables \ref{tab:southern_tablelands} and \ref{tab:snowy_mountains}). The best prediction results were $r=0.76$ and $\text{RMSE}=5.89$ for Southern Tablelands region, and $r=0.68$ and $\text{RMSE}=5.95$ for the Snowy Mountains region.

(2) The correlation between DESIS hyperspectral data and plant species richness was region-specific. Modelling the correlation separately for each region produced better results than building a single model for all regions.

(3) The relative importance analysis conducted among DESIS bands showed that the red-edge, red, and blue spectral regions are more important in predicting plant species richness than the green bands and the near-infrared bands beyond red-edge (noting that the SWIR region is not sampled by DESIS).

(4) The DESIS hyperspectral data performed better than multispectral data in predicting plant species richness, indicating that the provision of richer information in the spectral domain is important for diversity mapping.

Results shown in this study provided a quantitative reference on the potential for spaceborne hyperspectral data to be used in the mapping of on-ground plant species richness. Future studies should focus on extending the current approach to larger areas, investigating the potential of upcoming hyperspectral missions that extend into the SWIR region, and exploring the combination of data from other sensor types.

\section{Acknowledgement}

The authors are grateful to the anonymous reviewers for their important and insightful comments for improving this manuscript.

\bibliography{ref}

\end{document}